\documentclass{ieeeaccess}

\usepackage{cite}
\usepackage{amsmath,amssymb,amsfonts}
\usepackage{algorithmic}
\usepackage{graphicx}
\usepackage{textcomp}
\usepackage{url}
\usepackage{balance}
\usepackage{epstopdf}	
\usepackage{caption}
\usepackage{subfigure}

\graphicspath{{figures/}}

\def\BibTeX{{\rm B\kern-.05em{\sc i\kern-.025em b}\kern-.08em
    T\kern-.1667em\lower.7ex\hbox{E}\kern-.125emX}}

\begin{document}


\title{\LARGE \bf
Autonomous Execution of Cinematographic Shots with Multiple Drones}

\author{\uppercase{Alfonso Alc\'{a}ntara}\authorrefmark{1}, \uppercase{Jes\'{u}s Capit\'{a}n}\authorrefmark{1}, \uppercase{Arturo Torres-Gonz\'{a}lez}\authorrefmark{1},  
\uppercase{Rita Cunha\authorrefmark{2}, and An\'{i}bal Ollero}\authorrefmark{1}}
\address[1]{GRVC Robotics Laboratory, University of Seville, Spain (e-mail: [aamarin,jcapitan,arturotorres,aollero]@us.es)}
\address[2]{Instituto Superior Tecnico, Lisbon, Portugal (e-mail: rita@isr.tecnico.ulisboa.pt)}
\tfootnote{
This work was partially funded by the European Union's Horizon 2020 research and innovation programme under grant agreements No 731667 (MULTIDRONE), and by the MULTICOP project (Junta de Andalucia, FEDER Programme, US-1265072).}

\markboth
{A. Alcantara \headeretal: Autonomous Execution of Cinematographic Shots with Multiple Drones}
{A. Alcantara \headeretal: Autonomous Execution of Cinematographic Shots with Multiple Drones}

\corresp{Corresponding author: Jesus Capitan (e-mail: jcapitan@us.es).}

\begin{abstract}
This paper presents a system for the execution of autonomous cinematography missions with a team of drones. The system allows media directors to design missions involving different types of shots with one or multiple cameras, running sequentially or concurrently. We introduce the complete architecture, which includes components for mission design, planning and execution. Then, we focus on the components related to autonomous mission execution. 
First, we propose a novel parametric description for shots, considering different types of camera motion and tracked targets; and we use it to implement a set of canonical shots. 
Second, for multi-drone shot execution, we propose distributed schedulers that activate different shot controllers on board the drones. Moreover, an event-based mechanism is used to synchronize shot execution among the drones and to account for inaccuracies during shot planning. 
Finally, we showcase the system with field experiments filming sport activities, including a real regatta event. We report on system integration and lessons learnt during our experimental campaigns.  
\end{abstract}


\titlepgskip=-15pt

\maketitle

\section{Introduction}
\textit{Unmanned Aerial Vehicles} (UAVs) or drones are becoming mainstream for imagery and cinematography, mainly due to their maneuverability and capacity to produce unique shots in comparison with static cameras and dollies. 
The use of teams with multiple drones broadens the spectrum of artistic possibilities for media production, as several action points could be filmed concurrently or alternative perspectives applied to the same subject. This is even more accentuated in outdoor settings, where drones may need to cover large-scale scenarios with multiple action points.
Nowadays, the market offers many commercial platforms for both amateur and professional cinematographers. Nonetheless, operating these systems is complex and usually requires two expert pilots per drone; one controlling the drone and another for the camera. The task of synchronizing manually drone and camera motion while ensuring safety and aesthetic video outputs remains challenging, and hence, pilots get overloaded.
\begin{figure}
    \centering
    \includegraphics[width=0.9\columnwidth]{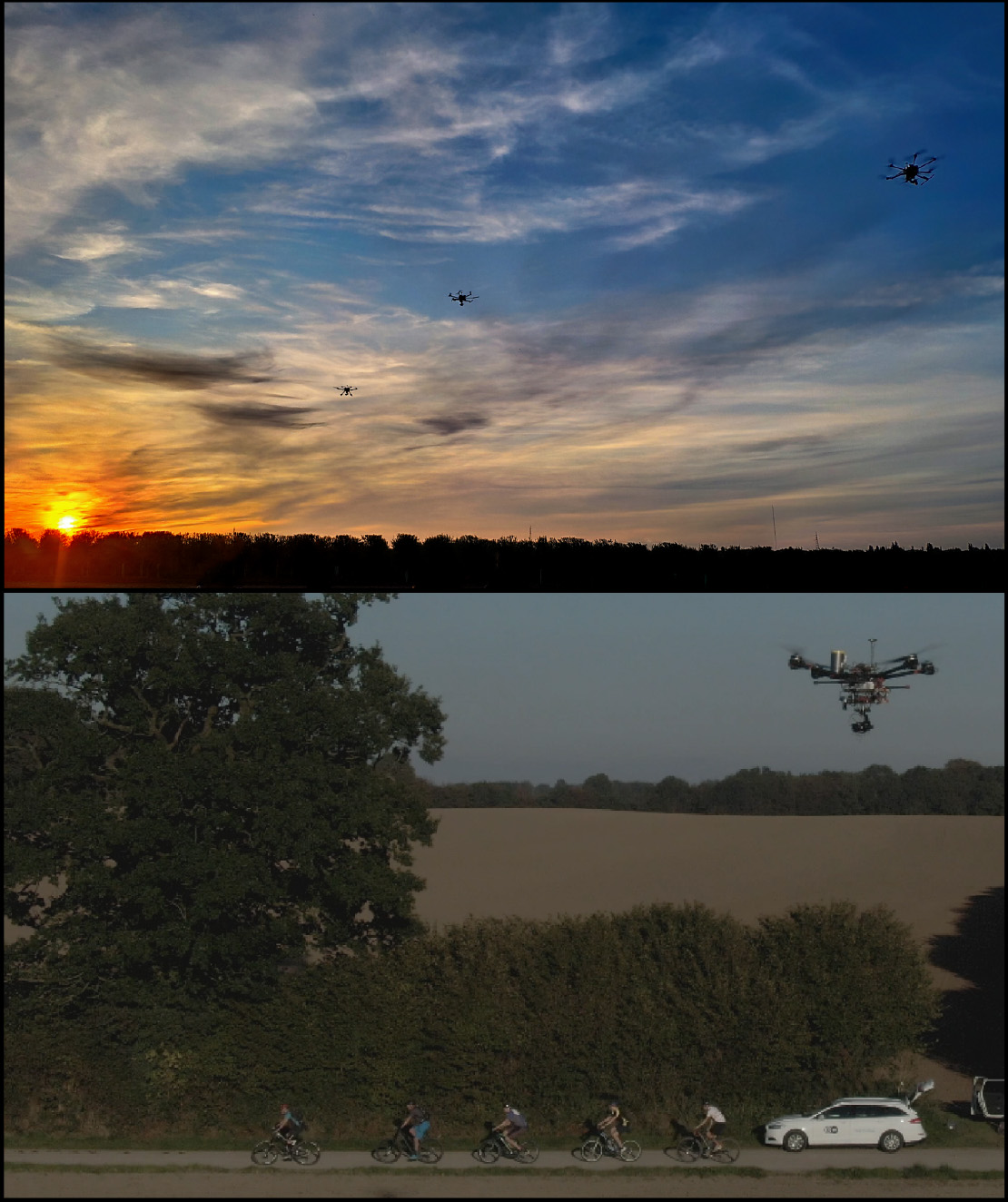}
    \caption{Different views of our mock-up experiments with multiple drones filming sport activities.}
    \label{fig:image_intro}
\end{figure}
Certainly, there exist several commercial products (e.g., \textit{DJI Mavic}~\cite{mavic} or \textit{Skydio}~\cite{skydio}) that alleviate the aforementioned complexity by implementing partially autonomous functionalities. They typically provide \textit{auto-follow} features to identify and track an actor visually or with GPS, as well as simplistic collision avoidance. However, they do not consider high-level cinematographic principles for shot performance nor multi-drone teams, and only implement a reduced set of shots. Therefore, there is still a need for autonomous systems that are intelligent enough to execute cinematography shots with multiple action points and multiple drones. This implies, for instance, predicting how the scene will evolve and being able to schedule shots that may be happening sequentially or in parallel, as well as coping with possible failures and contingencies. 

Recently, the EU-funded project \textit{MultiDrone}\footnote{https://multidrone.eu}, where our work is framed, has finished successfully; producing an integrated system for autonomous cinematography with multi-drone teams in outdoor sport events (see Figure~\ref{fig:image_intro}). The project covered all aspects in the complete system: a set of high-level tools so that the media end-user defines all the shots that compound the mission; planning algorithms to assign and schedule shots among the drones efficiently; and methods to execute those shots autonomously with the drones in a distributed manner.           
In this paper, we present the general architecture of the system and focus on the last part devoted to shot execution. 
In particular, we introduce our distributed system for autonomous execution of cinematographic shots with multiple drones. First, we select a set of canonical shots from the cinematography literature, and define the required properties to describe them by means of autonomous controllers. Then, we devise a system that allows the multi-drone team to execute concurrent shots in a distributed manner, by means of synchronization events. 

In MultiDrone project, we proposed a new taxonomy for cinematographic shots with drones~\cite{mademlis_tb19,mademlis_cs19}, and with the support of experts from the media production companies involved in the project, we selected a set of representative shots to be implemented autonomously by the system. These shots can be defined by the \textit{media director} through a high-level graphical interface with a novel language that we created for cinematography mission description~\cite{montes_appsci20}. The director indicates desired shot types, starting times/positions and durations; but she/he does not assign specific drone cinematographers to them. Instead, the system computes autonomously feasible plans for the drones~\cite{caraballo_arxiv20}, considering constraints such as their remaining battery, no-fly zones, collision avoidance, etc. Each drone gets scheduled one or several shots, together with the \textit{events} that will trigger each shot. These shots may be sequential, filming different action points along time (or the same with different views); or they may be concurrent shots with multiple drones filming one or several action points. The focus of this paper is on mission execution, so we assume these planned schedules for each drone as a starting point. Different planning techniques may be used to compute those  schedules~\cite{torres_robot17,caraballo_arxiv20}. 

\subsection{Contributions}

This work presents our multi-drone system for autonomous execution of cinematography missions. We introduce the architecture of the complete system and then describe the components related with shot execution. For that, we propose a distributed scheduler that runs on board each drone and activates different shot controllers depending on the shot type. These controllers are in charge of both drone and gimbal motion. Then, an event-based system is used to synchronize shot execution among the drones and ensure proper coordination. Furthermore, we increase the system robustness by considering contingency plans. In particular, our system is able to react to possible drone failures (e.g., lack of battery or GPS signal), re-planning the remaining shots with the available drones and letting the failed ones to perform emergency maneuvers.

In Section~\ref{sec:soa}, we review the state of the art for drone cinematography. Then, our main contributions with respect to previous works are the following:

\begin{itemize}
    \item We present a complete architecture for autonomous execution of cinematography missions with a team of drones (Section~\ref{sec:architecture}). We formulate the problem of autonomous cinematography as two steps: mission planning and execution. In this paper, we describe our novel solution for mission execution, integrating components for target tracking, drone motion and gimbal control. Even though we implement a representative set of canonical shots, we also generalize the way to describe parametric shots, making the system easily extensible. In this sense, we allow for different camera motion modes, including actual target tracking and predefined virtual rails.     
    
    \item We describe our method for cinematography mission execution (Section~\ref{sec:distributed}), which is agnostic to the planner used to schedule and assign shots to the available drones. We propose distributed schedulers that trigger the execution of the different shots based on starting events that may be generated manually or autonomously (e.g., a certain actor reaching an action point). This works as a synchronizing mechanism for multi-drone shots but also makes the system robust to uncertainties in the planning phase (e.g., the planned starting time of some action getting delayed). Moreover, our onboard controllers implement shots autonomously decoupling gimbal and drone motion, which improves robustness to noisy actor measurements (compensating with gimbal control). 
    
    
    \item We provide an open-source implementation of our system using off-the-shelf hardware, and validate it for outdoor media production with multiple drones (Section~\ref{sec:experiments}). 
    In particular, we show our field experiments filming several sport activities (including a real regatta), with the system running all components onboard in real time. We also report on lessons learnt after our experimental campaigns within the framework of the MultiDrone project, which are backed up by the feedback provided by the media experts involved in the project. 

\end{itemize}
\newpage
\section{Related Work}
\label{sec:soa}


\textit{Commercial products:} There are multiple commercial products for drone cinematography in outdoor settings. On the one hand, aerial platforms like \textit{DJI Mavic}~\cite{mavic}, \textit{Skydio}~\cite{skydio}, \textit{3DR SOLO}~\cite{3dr} or \textit{Yuneec Typhoon}~\cite{yuneec} offer good performance, including some semi-autonomous functionalities for tracking moving targets visually or by GPS, as well as simplistic collision avoidance. However, the set of shots is predefined and not easily extensible, as their software suites are not open-source. Besides, they do not consider multi-drone systems nor multi-shot scheduling. On the other hand, there are commercial applications to enhance the user experience. \textit{Skywand}~\cite{Skywand} is a virtual reality system that allows the user to explore the scene and select desired key-frames within the virtual environment. Then, the system computes a drone trajectory for a smooth shot containing these key-frames. \textit{Freeskies CoPilot}~\cite{FreeSkies} is a mobile software suite that offers similar functionality but with a simple 3D map instead of a virtual reality interface. In both cases, the resulting drone autonomy and environment perception are minimal, the cinematography plans consist of example key-frames and they cannot be adjusted online.

\textit{Autonomous systems with one drone:} In the robotics literature, there are works that propose partial autonomy but not complete integrated systems. For instance, PID~\cite{teuliere_iros11} or LQR~\cite{naseer_iros13} controllers have been considered for target tracking but without considering cinematographic rules. 
In~\cite{kang_ral18}, a system to support operators with certain autonomy is presented. Simple touch human gestures on a screen are interpreted in order to be translated into drone and gimbal movements.
In~\cite{coaguila_flairs16}, a discrete probabilistic decision-maker is used to take frontal shots of a moving target. They select between two actions: staying or moving to a new goal location facing the target. The idea is to estimate target's intentions (changing location/orientation or staying) and minimize the camera movement accordingly.

More recently, some works have considered cinematographic principles more explicitly when filming dynamic targets with a single drone in outdoor scenarios. For instance, a real-time dynamic camera planning strategy based on limbs movement detection is presented in~\cite{Huang}. Visual tracking is also performed in~\cite{bonatti_iros19}, where camera motion is planned addressing collision avoidance and aesthetic constraints. 
The same authors have proposed a novel method based on reinforcement learning~\cite{gschwindt_iros19} to achieve visually pleasant shots. In a similar line, \cite{huang_icra19} implements an algorithm to imitate (learning from demonstration) professional cameraman's intentions for capturing aerial footage of a single subject. 

Close to our work, a complete system for drone cinematography in unstructured environments is presented in~\cite{bonatti_jfr20}. They combine vision-based target tracking with a real-time motion planner that avoids collisions and fulfills artistic guidelines. They show impressive field experiments, but their focus is mainly on mapping and obstacle avoidance rather than multi-shot scheduling. Moreover, only a single drone is considered, as well as a simplified set of shots: left, right, front, back.

\textit{Virtual camera control for cinematography:} Designing smooth trajectories for virtual cameras using cinematographic techniques has been widely studied in computer animation. A complete review can be found in~\cite{Christie2008}. The common idea is to formulate some kind of offline optimization problem in order to generate smooth camera trajectories that satisfy aesthetic and cinematographic constraints. For example, there exist specific tools to support the planning of aerial shots in 3D virtual environments~\cite{joubert_siggraph15,gebhardt_chi16}. The user specifies 3D positions and a timed reference trajectory is generated for the camera. Even though these trajectories are optimal in terms of aesthetic objectives, physical feasibility considering drone dynamics is not always ensured. 
In~\cite{joubert_siggraph15}, violations of these dynamic constraints in the planned trajectories are at least detected, and the velocity along the trajectory can be adjusted by the user at execution time. A similar application for outdoor filming design is proposed in~\cite{joubert_arxiv16}, and the timing for the shots is considered by means of easing curves that drive the drone along the planned trajectory (i.e., the curve can modify its velocity profile). In~\cite{gebhardt_chi16}, an iterative quadratic optimization problem is formulated to obtain smooth trajectories for the camera and the look-at point (i.e., the place where the camera is pointing at). Collision avoidance constraints are included, but the method is only demonstrated indoors. Alternatively, other works try to reduce the search space of the optimization problem to achieve real-time performance by planning in a \emph{toric space}~\cite{lino_tog15} or interpolating polynomial curves~\cite{galvane_eurographics16, joubert_arxiv16}. 
In general, many of these methods related to computer graphics assume full knowledge of the scenario and they do not cope with the constraints involved in real drone platforms. Moreover, those implemented outdoors, do not consider moving targets and are limited to static or close-to-static guided tour scenes. 

\textit{Autonomous systems with multiple drones:} Regarding the use of multiples drones, some applications related to cinematography are worth mentioning. For instance, the authors in~\cite{Petracek2020} propose a multi-drone system for documentation of historical buildings. While one of the drones is taking pictures, the others maintain a formation to illuminate the scene adequately. A multi-drone system for target localization outdoors is presented in~\cite{price_ral18}. They use \textit{Model Predictive Control} (MPC) for trajectory optimization, and tackle inter-drone avoidance with a technique based on potential fields. The work in~\cite{SAEED2017} proposes a method to place as few drones as possible to cover without occlusion all targets in a scenario. However, this is done in a 2D space and considering that cameras must always be facing the targets. Though related, none of these works are thought for cinematography in dynamic environments. 

More related to our work, an approach for cinematography with multiple drones is described in~\cite{naegeli_tg17}. They resolve a non-linear optimization to generate 3D trajectories for the drones. Aesthetic objectives and collision avoidance between the drones and with the filmed actors are considered. The problem is solved on each drone in a distributed fashion, after exchanging planned trajectories; and a receding horizon technique is used to achieve real-time performance. The authors extend their own previous work~\cite{nageli_ral17} by including multiple drones and preference trajectories from the user as virtual trails. Although the approach is quite promising for autonomous cinematography, it is only tested at indoor settings and does not consider the scheduling of multiple shots, as we do.

The work in~\cite{galvane_tg18} is quite close to ours, as the authors also propose a complete architecture for cinematography with multiple drones. They apply non-linear optimization in a novel drone toric space to produce polynomial trajectories that improve video quality. For that, they minimize curvature variation and integrate constraints for collision avoidance. The motion of the multiple drones around dynamic targets is coordinated by means of a master-slave approach that resolves conflicts: only one master drone is supposed to be shooting the scene at a time, while the slaves offer alternative viewpoints or act as replacements. Moreover, the user can only select among different framing types. Instead, our system adds more flexibility, as we define framing and shot types; as well as introduce multi-view shots more explicitly, allowing different types of shot to happen concurrently. Besides, the system in~\cite{galvane_tg18} is only tested at indoor settings, with a Vicon motion capture system that provides accurate positioning for all targets and drones.


\section{System Architecture}
\label{sec:architecture}

In this section, we present the complete architecture of our autonomous system for multi-drone cinematography. We assume that there is a \textit{media director} in charge of designing the mission by describing multiple shots from a high-level and artistic point of view. Then, this director is supported by autonomous components that are able to compute plans to perform the designed shots and execute the mission with a team of drone cinematographers. 
Our system separates the whole cinematography problem into two sub-tasks: mission planning and mission execution. 

\textit{Mission planning:} Given an input cinematography mission, this sub-task consists of deciding which drone should execute each of the shots. The director specifies for each shot (among other parameters) a starting position and time for the action to be filmed, as well as the desired duration and type. Taking into account the initial position and remaining flight time of the drones, a schedule with the shots assigned to each drone must be computed. 

This problem can be solved with scheduling and task allocation algorithms. Each shot represents a task with a duration and an estimated starting time and position; and it must be ensured that each drone has enough flight time to cover all its assigned shots. After every shot, a path to reach the starting position of the next shot is necessary. For that, an estimation of the ending position of the drone after the shot is required. For certain sports events like those in our work (e.g., rowing or cycling races), this is assumable, as targets move along a predefined route with an approximate known speed. We developed our own algorithm for optimal mission planning with time constraints and avoiding inter-drone conflicts~\cite{caraballo_arxiv20}, but our architecture could accommodate alternative methods~\cite{natalizio_tmc19}. Our algorithm maximizes the percentage of shots covered by the multi-drone team and it provides as output a list of actions for each drone in the team. We consider two types of actions: \textit{Navigation Actions} (without filming) to take off, land and navigate from one shot to the next one; and \textit{Shooting Actions} to execute a specific shot. Shooting Actions involve concurrent drone and gimbal control, and they can have a starting \textit{Event} associated which triggers execution. We leave mission planning out of scope of this paper and concentrate on the problem of mission execution.    

\textit{Mission execution:} Given a plan for a cinematography mission, i.e., the list of Shooting and Navigation Actions assigned to each drone, this sub-task consists of executing those shots in a synchronized manner with a multi-drone team. This means triggering gimbal and drone controllers that depend on the shot type, avoiding drone collisions and performing target tracking. 

We solve this problem by means of a set of distributed shot schedulers and executors that run on board the drones. For mission planning, we assume that the director can estimate the occurrence time for the Events triggering Shooting Actions. We also assume that target trajectories can be predicted approximately. However, the system tolerates errors in those estimations to a certain extent, since it reacts online during mission execution in two manners: (i) drones wait at shot starting positions before triggering execution, to account for delays on the actual action to be filmed; and (ii) drones can track actual target trajectories instead of planned ones during shot execution, to account for possible deviations.

\subsection{System Overview}
\label{sec:system}

\begin{figure*}[tb]
    \centering
    \includegraphics[width=0.9\textwidth]{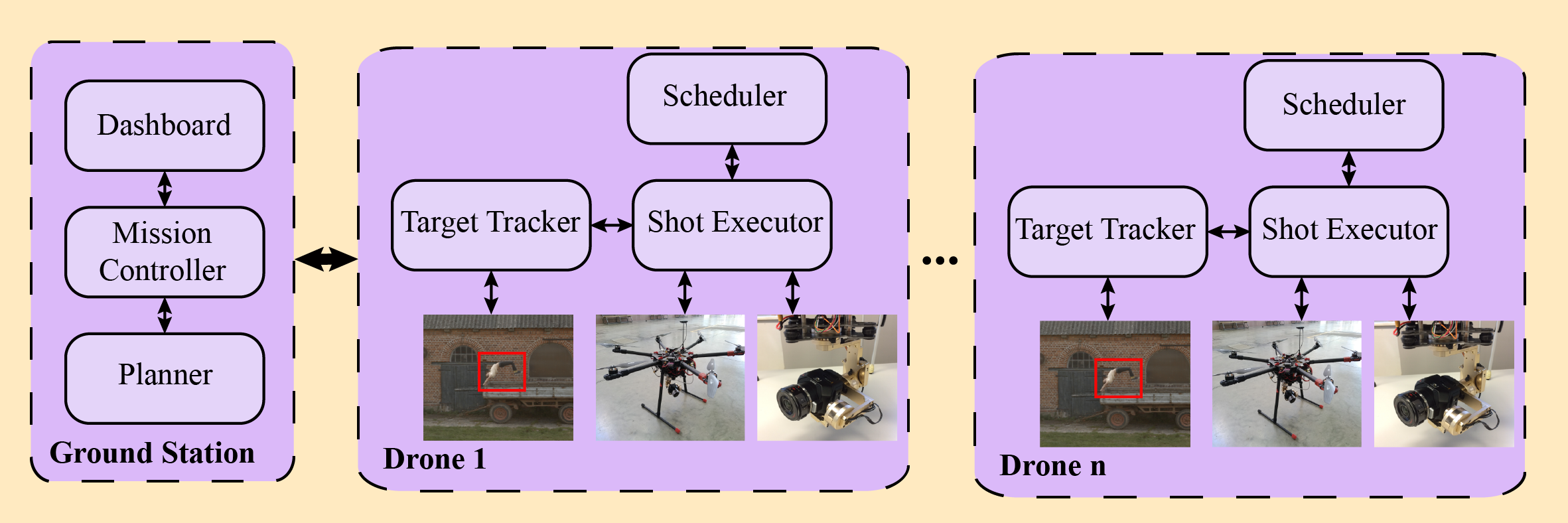}
    \caption{System architecture for multi-drone cinematography. Components for mission design and planning run on a Ground Station, but components for mission execution run on board the drones.}
    \label{fig:system}
\end{figure*}

Figure~\ref{fig:system} shows the complete architecture of our system. Components related to mission planning are executed on a \textit{Ground Station} that interfaces with the director, whereas components related to mission execution run mainly on board the drones. The \textit{Dashboard} is a graphical tool for human-computer interaction between media end-users and the rest of the system. This component allows the director to design cinematography missions, including all shot descriptions and their triggering Events, when needed. 
For instance, a director could design a mission to film a rowing race; and specify a \textit{lateral} shot from the \texttt{START\_RACE} Event to the end of the race, and an \textit{orbital} shot starting with the \texttt{FINISH\_LINE} Event, i.e., when the boats reach the finish line. 
We proposed a novel cinematography language~\cite{montes_appsci20} so that the director's input is written with a specific syntax that is later understandable for our planning components. 

On the Ground Station, there is another central component called \textit{Mission Controller}, which manages the whole planning and execution process for a mission. This module receives director's input through the Dashboard and it uses the \textit{Planner} component to compute feasible plans in order to execute the mission. Then, the Mission Controller sends to each drone its plan, which basically consists of a list of Shooting Actions to execute assigned shots, with interleaved Navigation Actions to fly between shots. During mission execution, the Mission Controller monitors drone status for possible contingencies and sends out the triggering Events as they actually occur. Depending on the Event, this occurrence may be detected automatically by the Mission Controller or indicated manually by the director. 

Components on board the drones manage mission execution. Communication with the Ground Station is done by means of an LTE link~\cite{mademlis_eusipco19} through the \textit{Scheduler} components, which are the ones receiving plans and Events from the Mission Controller. They are in charge of executing shots in a distributed manner with multiple drones. Each Scheduler listens to Events and starts/stops the execution of Shooting Actions as required. These Events act as a synchronizing mechanism for multi-camera shots, since all involved drones wait for the same Event to start. 
Shooting Actions are carried out by calling the \textit{Shot Executor} component, which implements drone and gimbal controllers. Depending on the shot parameters, the Shot Executor adapts its controllers to perform the corresponding shot. A \textit{Target Tracker} module is necessary to provide positioning of the target, which is used by the Shot Executor to point the gimbal and move the drone accordingly. Navigation Actions are also managed by the Shot Executor, but with different controllers that do not consider gimbal motion nor cinematographic constraints. 

Our system is flexible to adapt to upcoming situations during execution. In particular, we allow for mission re-planning due to a director's choice or in case of contingencies. The former is triggered manually, but the latter is managed autonomously as follows. Schedulers report back to the Mission Controller the status of the mission execution, i.e., which action is each drone executing or waiting for. In case of an emergency in a drone, e.g. low battery or loss of GPS, the corresponding Scheduler is able to trigger an emergency maneuver (landing safely), but at the same time, it informs the Ground Station about the situation. Then, the Mission Controller starts a re-planning procedure through the Planner component, considering only the available drones and the remaining shots to execute. Once those new plans are sent to the drones, each of them will finish with its ongoing action, and will append the new list of actions behind. The other safety mechanism that is considered in our architecture is collision avoidance. This is integrated at planning level within the Planner, and at execution level within the Shot Executor. First, our Planner uses a high-level map of the environment (including no-fly zones due to obstacles, audience, etc) to provide collision-free paths. It also resolves inter-drone conflicts when their paths go too close. Second, our Shot Executor runs collision avoidance online to react to unexpected situations and keep inter-drone safety distances.  

\subsection{Shot Description}
\label{sec:shotDescription}

Our multi-drone system performs autonomously a series of shots that are represented by Shooting Actions. All properties for each shot are encoded through the attributes of its corresponding Shooting Action. Table~\ref{tab:sa} depicts the definition of a shot, with multiple properties that can be specified when designing the shot.   

\begin{table}[htb]
\caption{Attributes of a Shooting Action for shot definition.}
\begin{center}
\begin{tabular}{p{0.15\columnwidth}p{0.20\columnwidth}p{0.5\columnwidth}}
	\hline
	\hline
	\textbf{Attribute} & \textbf{Data type} & \textbf{Description}\\ \hline  
	Shot type & Discrete value & Chase, lateral, orbit, etc.\\ 
	Framing type & Discrete value & Long shot, medium shot, close-up shot, etc.\\ 
	Start Event & String & Event that triggers this action \\ 
	Duration & Time & Duration of the shot \\ 
	RT\newline path & List of global positions & Estimated path of the RT \\ 
	RT speed & Float & Speed along the RT path \\ 
	RT mode & Discrete value & virtual-traj, virtual-path\newline or actual-target \\ 
	RT ID & String & Identifier of the RT to follow \\ 
	ST type & Discrete value & Virtual, real or none \\ 
	ST ID & String & Identifier of the ST to follow\\ 
	Shooting\newline parameters & Set of\newline parameters & E.g., relative distance to RT, angular velocity in an orbit, etc.\\ \hline
	
\end{tabular}
\label{tab:sa}
\end{center}
\end{table}

The \textit{shot type} describes the kind of movement of the camera with respect to the action, i.e., chasing, orbiting around, etc. We will define in Section~\ref{sec:shots} all shot types, together with the \textit{shooting parameters} defining their geometry. Apart from the shot type, we need to specify the \textit{framing type} (i.e., how close the action will appear on the image), the \textit{duration} and the \textit{starting Event}. This latter is optional, if not specified, the shot would start right after the previous one. Besides, we create two relevant concepts to describe shots: the \textit{Reference Target} (RT) and the \textit{Shooting Target} (ST). The RT is used to guide drone motion, as the drone should follow this target describing its corresponding type of shot. The ST is used to guide gimbal motion, as the camera should point at this target when filming. Both targets could coincide, but not necessarily. For instance, we may want a camera moving along a lateral rail but filming a static scene or an actor moving in a different direction. 

We specify the \textit{RT path} as a list of waypoints expressed in global coordinates, and depending on the \textit{RT mode}, we define three different kinds of motion for the drone:

\begin{itemize}
\item \textit{Mode virtual-traj}: A virtual drone trajectory is specified. The drone should move along the rail indicated in the RT path and at the velocity specified in \textit{RT speed}.
\item \textit{Mode virtual-path}: A virtual drone path is specified but no speed is provided. The drone should move along the rail indicated in the RT path but at the speed of an actual target, which would be indicated by the RT ID. 
\item \textit{Mode actual-target}: No virtual path is indicated for the drone, which should move following an actual target specified by the ST. 
\end{itemize}

The above modes widen the spectrum of possibilities for the director and were actually recommended by media experts from our end-user partners in the MultiDrone project.
On top of that, we can track different targets with the drone and on the image, i.e., having non-coincident RT and ST. We consider three types of ST: (i) \textit{virtual}, if it is specified as a virtual point or path, i.e., the RT path; (ii) \textit{real}, if it is an actual physical target (e.g., a cyclist, a runner, etc.) whose position can be estimated, for instance through visual detection or with a mounted GPS; and (iii) \textit{none}, if the camera is just fixed or following a predefined motion. In case of a real ST, an \textit{ST ID} can be indicated to identify the specific target to track visually or the corresponding GPS transmitter. A similar role plays the \textit{RT ID} when we use the virtual-path RT mode to track an actual target with the drone.

Finally, notice that our shot description does only require a starting Event for particular shots. The director may want to perform a series of sequential shots after a given Event, to take several views along the line of action. For that, she/he would only need to specify the starting Event for the first shot, and the others would happen consecutively. 
Furthermore, it is important to highlight how multi-drone shots are considered within this framework. The director could design multi-camera shots to be performed by a formation of multiple drones simultaneously. For that, she/he could assign the same starting Event and RT to several Shooting Actions. Thus, all drones involved would track together a common reference trajectory, implementing complementary shots of the same or different types. The shooting parameters for each Shooting Action would determine the geometry of the formation, and the starting Event would synchronize the motion so that they all start shooting simultaneously.

\subsection{Canonical Shots}
\label{sec:shots}

In this section, we describe the set of shots that have been implemented for our system. In the cinematography literature there is a lot of information about cinematographic rules and canonical types of shots~\cite{smith2016photographer}. Within the context of the MultiDrone project, we studied a wide spectrum of shots\cite{mademlis_tb19,mademlis_cs19}, and following the recommendations of the media experts in the project, we selected our canonical list of representative shots for the autonomous system.
In the following we describe shots types and the specific shooting parameters considered for each of them. Table~\ref{tab:params} summarizes all parameters.

\begin{table}[htb]
    \centering
    \caption{Shooting parameters for each shot type.}
    \label{tab:params}
    \begin{tabular}{ll}
        \hline\hline
        \textbf{Shot type} & \textbf{Shooting parameters} \\ \hline
        Static & $pan_s$, $tilt_s$, $pan_e$, $tilt_e$, $z_0$ \\
        Fly-through & $pan_s$, $tilt_s$, $pan_e$, $tilt_e$, $z_0$ \\
        Elevator & $z_s$, $z_e$ \\ 
        Chase/lead & $x_s$, $x_e$, $z_0$ \\
        Flyby & $x_s$, $x_e$, $y_0$, $z_0$ \\
        Lateral & $y_0$, $z_0$ \\
        Establish & $x_s$, $x_e$, $z_s$, $z_e$ \\
        Orbit & $r_0$, $azimuth_s$, $angular\_speed$, $z_0$ \\
        \hline
    \end{tabular}
\end{table}

\textit{Static:} The drone remains stationary above a fixed RT location, and this height is indicated by the parameter $z_0$. Since the RT represents a static position, the only RT mode that makes sense is virtual-traj. Depending on what the gimbal tracks, the ST type can be real or virtual. The ST type none can be used to implement shots scene-centered, in which the gimbal moves independently. In this case, the parameters $pan_s$, $pan_e$, $tilt_s$ and $tilt_e$ indicate the pan/tilt starting and ending angles, respectively.

\textit{Fly-through:} The drone flies through the scene following a predefined path with no specific target to track. As in the previous shot, the only possible RT mode is virtual-traj, as there is no actual target. The flight altitude over the RT path is indicated by the parameter $z_0$. The ST type is always none and there are extra parameters to describe gimbal movement along the shot duration: pan/tilt starting and ending angles ($pan_s$, $pan_e$, $tilt_s$ and $tilt_e$).

\textit{Elevator:} The drone moves vertically straight up or down tracking an actual target or a static position. The drone starts the shot above a given position (defined as the initial RT location) at altitude $z_s$, and it ends at $z_e$. Therefore, the RT mode is virtual-traj, but the ST type could be real or virtual.

\textit{Chase/lead:} The drone chases a target from behind with constant or decreasing distance; or leads it in the front with decreasing or constant distance. All RT modes are possible, depending on whether a virtual or actual target is followed; whereas only the real ST type makes sense. Regarding parameters, $z_0$ determines the drone height over the RT and $x_s$ and $x_e$, the starting and ending distances in the $X$ axis (pointing forwards) with respect to the RT.

\textit{Flyby:} The drone flies past a target normally overtaking the target as the camera tracks it. The RT could be virtual or real, so all RT modes are possible; whereas only the real ST type makes sense.
It needs as parameters distances with respect to the RT: $z_0$ for the altitude, $x_s$ and $x_e$ for the starting and ending distances in the $X$ axis, and the constant lateral distance $y_0$.

\textit{Lateral:} The drone flies beside a target with constant distance as the camera tracks it. The RT could be virtual or real, so all RT modes are possible; whereas only the real ST type makes sense.
It needs as parameters the $z_0$ altitude with respect to the RT, and the constant lateral distance $y_0$.

\textit{Establish:} The drone moves closer to a target from the front, typically with decreasing altitude. The RT could be virtual or real, so all RT modes are possible. The ST type could be real or virtual (e.g., to descend on a monument or static scene).
Both altitude and displacement in the $X$ axis with respect to the RT change during this shot, so it needs as parameters $z_s$, $z_e$, $x_s$ and $x_e$.

\textit{Orbit:} The drone moves around a target describing a full or partial orbit. The RT could be virtual or real, so all RT modes are possible. The ST type could be real or virtual (e.g., to orbit around a monument or static scene). The parameters in this case include the altitude over the RT ($z_0$), the radius of the circle ($r_0$), the starting azimuth angle in the orbit ($azimuth_s$) and the angular speed ($angular\_speed$).


\section{Distributed Mission Execution}
\label{sec:distributed}

In this section, we describe our autonomous components for cinematography mission execution. More specifically, this is the part of our system architecture that runs on board each of the drones. 

\subsection{Scheduler}

\begin{figure}
    \centering
    \includegraphics[width=0.8\columnwidth]{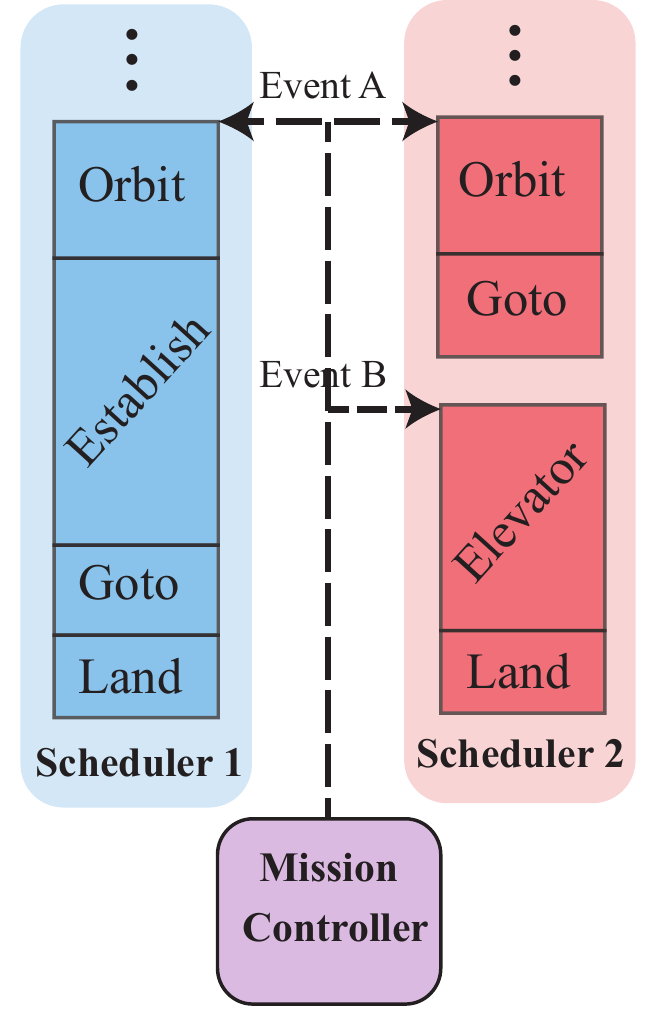}
    \caption{Example of the event-based procedure for distributed execution of a mission with two drones.}
    \label{fig:distributed_execution}
\end{figure}

The execution of drone shots is carried out by means of a distributed scheduling procedure. Each drone runs onboard a Scheduler component that receives the plan for that drone and coordinates the execution of the shots, with other drones involved and with respect to the actual development of the scene. In particular, the Scheduler receives a list of sequential Navigation and Shooting Actions. Navigation Actions only imply drone movement through the scenario, without filming. This is mainly to get to the starting position of a coming shot or to go for landing, so we only consider three types: \textit{take-off}, \textit{land} and \textit{go to waypoint}. In the last case, either a single waypoint or a list of waypoints to navigate through can be provided. Shooting Actions, instead, are those involving some filming of the scene. They require a special controller to take care simultaneously of drone and gimbal motion while a particular shot is executed. Thus, the set of available Shooting Actions coincides with the shots described in Section~\ref{sec:shots}.

The Scheduler controls the start and end of each action, handling the Shot Executor accordingly. 
For each Shooting Action, the drone is sent to its corresponding starting position, through a sequence of Navigation Actions that were computed by the Planner. Then, it keeps hovering at that starting position waiting for the Event associated with the Shooting Action. Once the Event arrives from the Mission Controller, the Scheduler activates the Shot Executor to start the Shooting Action. These Events represent actual action points of the scene being filmed, such a the start of a race, the runners reaching a particularly interesting point or the finish line. It is typical that the director wants to assign pre-designed sequences of shots for those moments. Moreover, if the Shooting Action has a specified duration, the Scheduler is in charge of waiting for that time before calling off the shot and continuing with the next action. In case of Navigation Actions, the Scheduler just waits for the notification of completion, and then it goes for the next action in the sequence. 


This event-based mechanism allows us to account for inaccuracies in the planning phase and for required adjustments during the actual filming of the scene. 
We assume that the Planner can estimate the occurrence time for the Events and an approximate target trajectory, what permits a plan computation. However, the system does not rely on estimated times for mission execution, but on the actual occurrence of the Events. Thus, we plan so that drones arrive \textit{earlier} than expected at their starting positions, and then wait for Events; considering possible delays in the actual scene being filmed. These Events could be detected online by the system in an automatic fashion. For instance, in rowing races, the launch signal can be communicated to the Mission Controller, and the race reaching specific points of the route can be detected by monitoring GPS trackers on board some of the boats. We also allow the director to send out Events manually to decide on shot triggering. Moreover, for multi-camera shots, the Events also act as multi-drone synchronizing signals. All involved drones will be waiting at their starting positions and the Event will ensure that they all start at the right time in parallel.      
Figure~\ref{fig:distributed_execution} shows an example of how the distributed scheduling works. In the example, two drones take an orbit shot in a synchronous manner, being triggered by a certain Event A in the scene. Then, right after the orbit, Drone 1 performs a establish shot and go back to its station to land; while Drone 2 goes to a new starting location to wait for Event B, which triggers an elevator shot that ends its mission.  

Additionally, the Scheduler component integrates a functionality for emergency management, which is crucial for safety. Each Scheduler monitors the drone status, being aware of hardware issues. In particular, we implemented low battery alerts and loss of GPS signal, but any other kind of contingency could be monitored. In case of failure, the Scheduler reports that status to the Mission Controller on the Ground Station. Then, the Mission Controller may decide to launch a re-planning procedure without the affected drone, reassigning its pending tasks to others.  
Simultaneously, the Scheduler carries out an emergency  maneuver. It cancels the action being executed and commands the drone to navigate to the closest base station for landing. For that, the Scheduler can plan safe paths that avoid no-fly zones. We implemented an off-the-shelf A$^*$ heuristic planner on a KML-based map that includes information about the positions of the base stations and the no-fly zones (areas with known obstacles or people gathering as audience).

\subsection{Shot Executor}


This component is in charge of executing Navigation and Shooting Actions. In order to execute a Shooting Action, we need to generate the desired trajectory, which is derived using the type shot, the shooting parameters and the target position. Given the target, which may be real or virtual depending on the RT mode, we make the drone behave as a trailer attached to that target~\cite{PEREIRA201777}. This method provides smooth reference trajectories to be tracked by the drone. This is particularly relevant when the target trajectory is very noisy (e.g., when following a real target) or defined by waypoints (e.g., as a virtual RT path). At the same time, by generating a trailer trajectory, a reference frame tangent to the path is obtained, which can be directly used to define the relative displacements encoded in the shooting parameters of each shot type, as well as the desired heading for the drone. For example, a chase shot would have the following parameters relative to this trailer reference frame: the constant altitude $z_0$, and the starting and ending distances to the target on the $X$ axis, $x_s$ and $x_e$.

Having the desired trajectory and an estimation of the current drone state, errors between current and desired position and yaw angle are used to generate velocity commands, applying a simple saturated proportional controller with a feedforward velocity term. These velocity commands are sent to the drone autopilot by means of our software library UAL (\textit{UAV Abstraction Layer})~\cite{real_ijars20}. This is a middleware abstraction layer that we developed to abstract drone navigation algorithms from the specific hardware details and interfaces of each autopilot. Thus, UAL provides a common interface to receive drone state, including positioning and battery level information. It also allows us to send velocity and position commands, as well as take-off and landing maneuvers. Regarding Navigation Actions, the Shot Executor performs them using directly our UAL interface, as no smooth trajectories are required for those actions. 

Apart from drone control, the Shot Executor is also in charge of controlling the gimbal to point the camera to the specified ST. If the ST type is none, the controller executes a predefined pan and tilt movement; if the ST type is real or virtual, the controller tracks a target.
Our gimbal has an IMU and a low-level controller that receives angular rate commands, defined with respect to the inertial frame. Therefore, the Shot Executor computes the desired angular motion independently from the drone movement and relative to the inertial reference frame. This desired angular velocity is computed using an attitude controller based on the error between the current and desired rotation matrices, incorporating both proportional and integral actions. More specific details about the mathematical formulation can be seen in~\cite{cunha_eusipco19}. Moreover, in order to compute this error matrix, the gimbal controller is able to accept both vision-based or GPS target measurements. In the first option, it receives as input 2D target positions on the image plane together with the direction of gravity in the body frame, which is encoded in the accelerometers measurements. Thus, no 3D information about the target position is required in that case. In the GPS-based mode, the desired camera orientation is computed by taking the difference between the drone and target positions (provided by their respective GPS receivers) to define a pointing direction and using the extra degree of freedom to enforce horizontal alignment.
In addition to drone and gimbal control, we also implemented some camera commands in the Shot Executor. In particular, the component is able to start and stop recording, autofocus, or modify some camera parameters, like zoom, ISO or white balance.

The Shot Executor requires an estimation of the target position that is provided by the Target Tracker component, which also runs on board the drone. This target positioning is needed whenever the drone is tracking an actual target, as RT or ST. We implemented two options for the Target Tracker in our system: (i) we used a GPS receiver on board the target together with a stochastic filter to estimate 3D target positions that were then sent to the drones; and (ii) we used a vision-based algorithm for target tracking that provided 2D positions on the image. The methods we applied for visual target tracking are based on light convolutional neural networks that can run on embedded computers in real time. This image processing part is out of the scope of this paper and further details can be seen in~\cite{nousi_icip19}. 

Additionally, the Shot Executor needs to take care of collision avoidance for safety reasons. For that, we used a reactive algorithm for collision avoidance in multi-drone teams~\cite{ferrera_sensors19}. In previous work, we developed that algorithm to resolve drone conflicts (i.e., possible collisions) with other teammates or external obstacles in a decentralized manner, applying roundabout maneuvers to avoid each other.    
We integrated this algorithm with our Shot Executor by running it as a reactive layer in parallel. This reactive layer sends warnings to the Shot Executor whenever a conflict is detected, together with velocity commands to resolve the conflict. Thus, the Shot Executor always prioritizes commands coming from the reactive layer over shot execution, in order to avoid collisions. Once the conflict warning disappears, shot execution resumes normally.

Finally, it is important to remark that our architecture allows us to use alternative solutions for the Shot Executor, as long as they address drone and gimbal control and implement the RT and ST concepts that we defined. Indeed, we also developed and tested within our architecture another algorithm for shot execution~\cite{ecmr19}. The algorithm plans optimal trajectories for the drone as it takes the shot, considering aesthetic aspects (e.g., generating smooth trajectories) and collision constraints. We tested this method running in real time on board the drones, achieving time horizons in the order of 10 seconds for trajectory planning.

\section{Field Experiments}
\label{sec:experiments}

We conducted extensive field tests to asses the performance of our complete system filming different outdoor activities. Since the system was developed for the MultiDrone project, our focus was on the sport use cases selected in the project, i.e., cycling/rowing races and parkour runners. The whole consortium devoted many efforts to integrate all software components into the team of aerial platforms developed in the project. In particular, we dedicated 9 weeks for physical integration throughout the last project's year, as well as 4 weeks for field tests with more than 40 hours of flight, split into two different campaigns in Germany and Spain. We setup several mock-up scenarios to recreate the aforementioned activities with amateur sportsmen, and we even filmed a real regatta event. In Germany, we used a field facility around a farm and next to a lake. The place is located in a village called Bothkamp, in the north of Germany, and it has permits to fly drones for amateur purposes. In Spain, we used another outdoor site in a farm 30 Km away from Seville.  

\subsection{System Integration}

\begin{figure}[t]
    \centering
    \includegraphics[width=0.8\columnwidth]{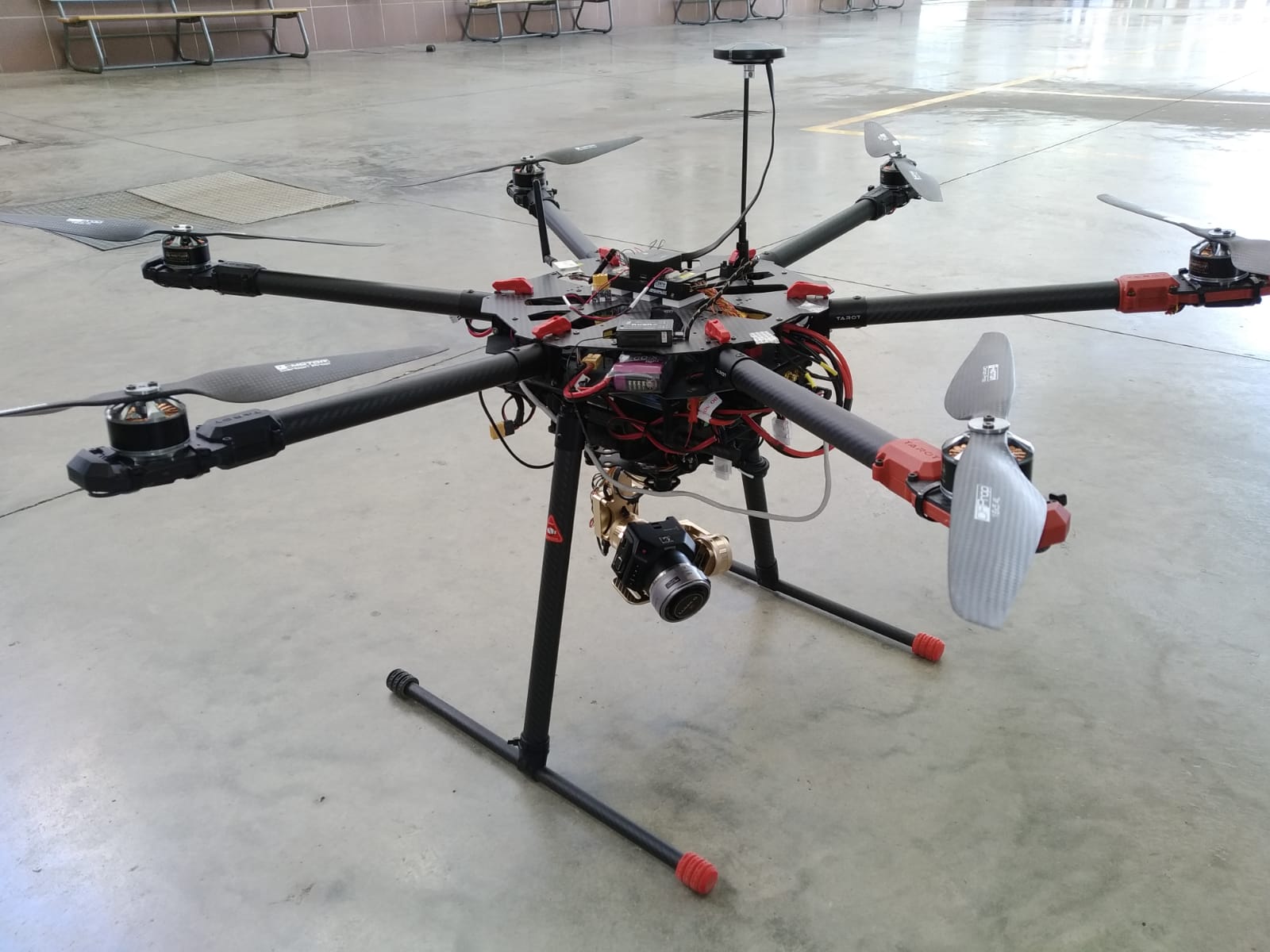}
    \caption{One of the drones used during our field experiments, with the cinematographic camera mounted on the gimbal.}
    \label{fig:drone}
\end{figure}

We used a simulation environment for early integration, and also so that the media director could double-check all missions before the actual shooting. Our simulation tool was based on \textit{Gazebo}~\cite{Koenig04designand} and the PX4~\cite{px4} SITL (\textit{Software In The Loop}) functionality for drone autopilots. We added a camera on a gimbal to the drones and interfaced them with our open-source~\footnote{https://github.com/grvcTeam/grvc-ual} UAL library~\cite{real_ijars20}, which abstracts users from the protocol details of each autopilot. 

The same software architecture ran in simulation and on our real drone platforms. We developed all software components as open-source in C++~\footnote{https://github.com/grvcTeam/multidrone\_planning}, using ROS Kinetic. We also mounted and integrated several drones like the one shown in Figure~\ref{fig:drone} for the experiments. They had the X6 frame from Tarot and were equipped with: a PixHawk 2 autopilot running PX4 for flight control; a RTK-GPS for precise localization; a 3-axis gimbal controlled by a BaseCam (AlexMos) controller receiving angle rate commands; a Blackmagic Micro Cinema camera; an Intel NUC i7 computer to run our software for drone execution; an NVIDIA TX2 computer dedicated to video streaming and image processing for target tracking; and a Thales LTE module to communicate with the Ground Station. We selected LTE to achieve better security and performance than WiFi in long-range distances.
Moreover, we devised a GPS target to be carried by selected human actors in some of the experiments. The device weighted around 400 grams and consisted of a RTK-GPS receiver with a Pixhawk controller, a radio link and a small battery. This target transmitted target 3D measurements to the Target Tracker on board the drones in real time (with a delay below 100~ms). The final 3D target estimation, after being filtered by the Target Tracker, was able to achieve centimeter level. These errors were compensated by our gimbal controller for tracking shooting targets on the video.

\subsection{Results}

\begin{figure}[tb]
    \centering
    \label{fig:parkour_scenario}

 
    \subfigure[Top view of the scene with virtual target and drone trajectories for each Shooting Action. In green, the parkour area.]{\label{fig:parkour_scheme} \includegraphics[width=1\columnwidth]{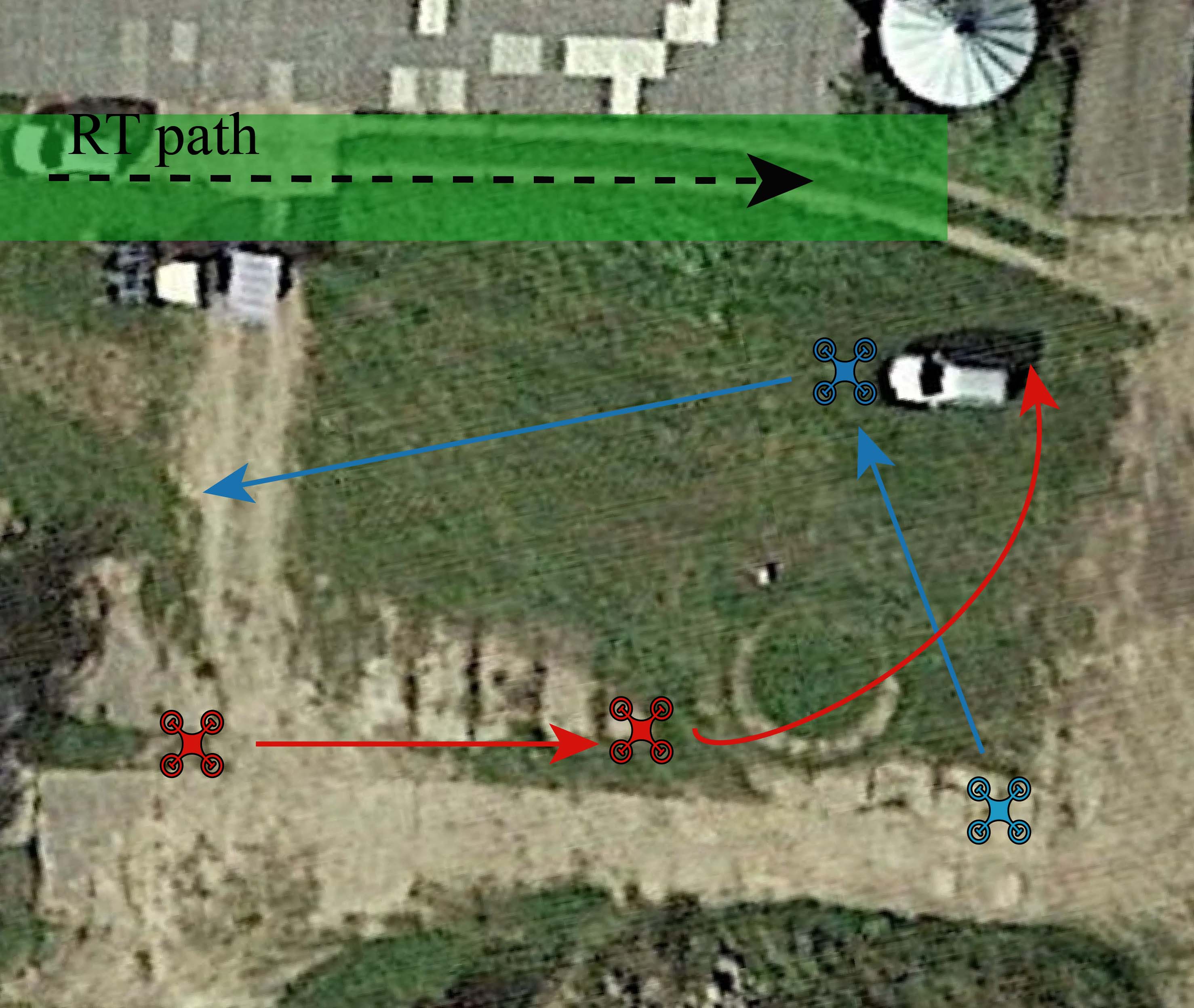}}
    \subfigure[Drone actions over a timeline. Both sequences of consecutive shots are executed in parallel triggered by the \texttt{START\_RACE} Event at time 20~s. A virtual target is tracked.]{ \label{fig:parkour_timeline}\includegraphics[width=1\columnwidth]{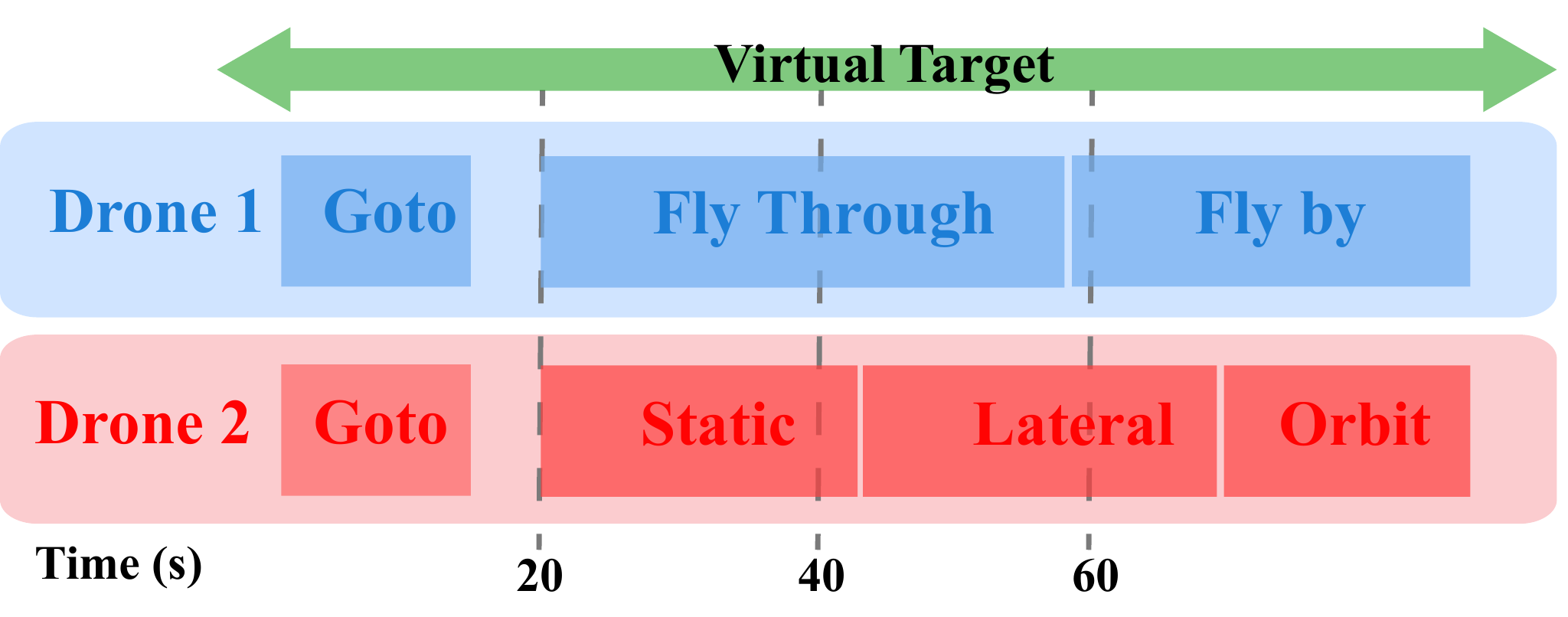}}
    \caption{Parkour mission with two drones and five different shots. Blue color corresponds to Drone 1 and red color to Drone 2.}
\end{figure}

\begin{figure}[tb]
    \centering
    \subfigure[Image from camera on board Drone 1 ]{\label{fig:parkour_mockup_1}\includegraphics[width=0.48\columnwidth]{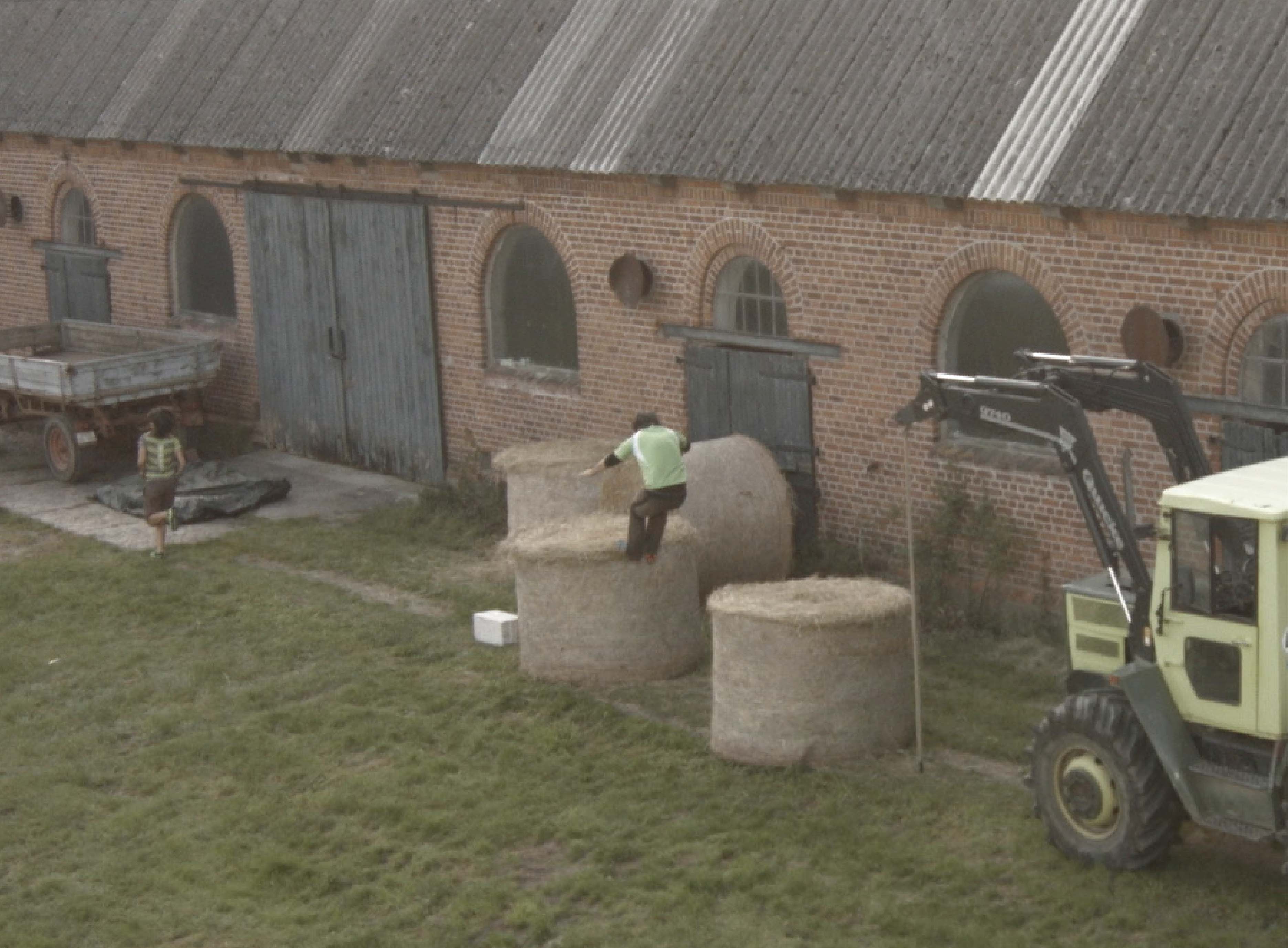}}
    \subfigure[Image from camera on board Drone 2]{ \label{fig:parkour_mockup_2}\includegraphics[width=0.48\columnwidth]{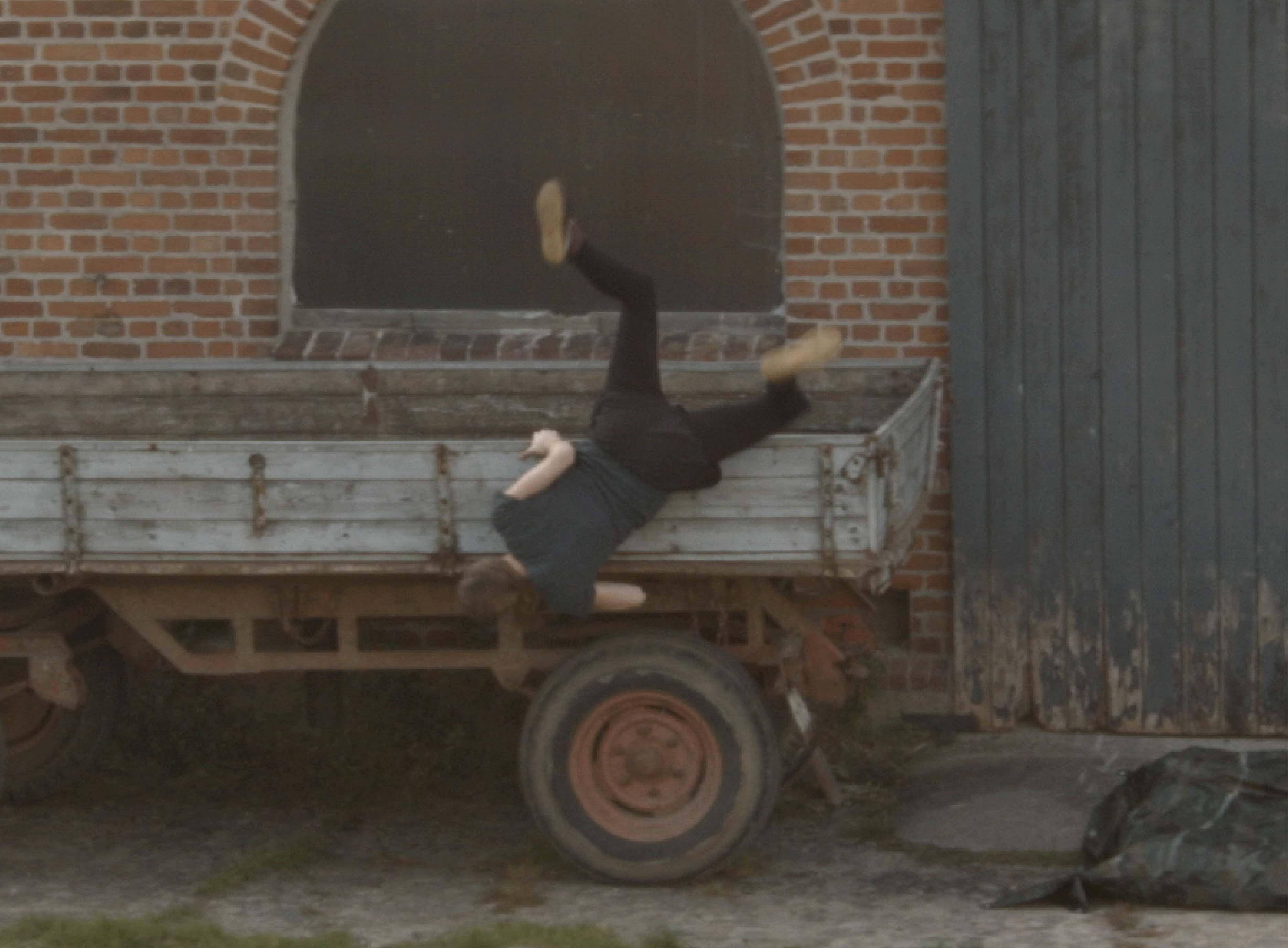}}
    \subfigure[Top view of the experiment. Both drones and one of the runners being filmed can be seen.]{\label{fig:general_mockup}\includegraphics[width=1\columnwidth]{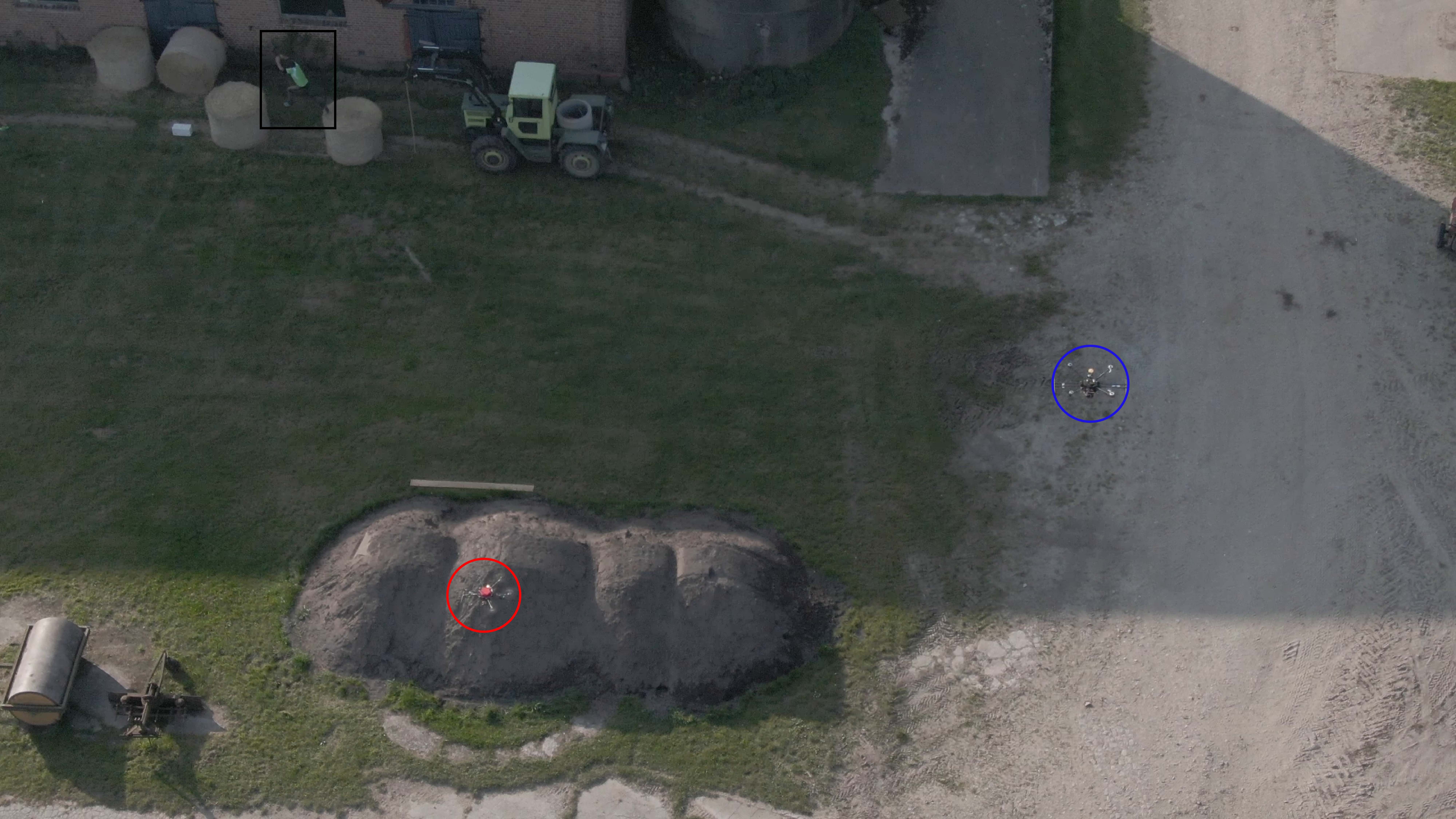}}
    \put(-85,57){ \textcolor{blue}{Drone 1}}
    \put(-180,23){ \textcolor{red}{Drone 2}}
    \put(-205,105){Target}
    \caption{Images from a mission with two drones filming a parkour activity.}\label{fig:parkour_mockup}
\end{figure}

In this section, we show example cinematography missions that followed the whole procedure through our architecture for autonomous filming: they were designed by a media expert with the Dashboard facility, and then planned and executed autonomously by the drones. Our main objectives are to demonstrate: (i) the integration of all the components working together; (ii) the feasibility of our system for autonomous cinematography with multiple drones outdoors; and (iii) the use of different shot types and RT / ST modes.

First, we illustrate parkour filming. Parkour is a sport activity where runners move freely over and through any terrain using only the abilities of the body, principally through running, jumping and climbing.
In our mock-up, we set up a specific longitudinal area with different obstacles and gathered a group of amateur parkourists to perform free-style maneuvers there. 
Figure~\ref{fig:parkour_scheme} depicts a scheme of a mission designed by our media director. Runners moved in the \textit{parkour zone} from left to right and the director designed a mission with 5 different shots. First, a sequence of a fly-through shot followed by flyby, triggered by the \texttt{START\_RACE} Event. Second, a sequence of a static shot, a lateral and an orbit, also triggered by the same \texttt{START\_RACE} Event. Since runners were moving freely in the scene, instead of tracking a particular one, the virtual-traj RT mode was used to specify a virtual trajectory for the RT in the parkour area. This RT path was used by the lateral, the flyby and the orbital shot, while the others had their own RT path independent of the runners. The ST was none for the static and the fly-through shots, and configured as virtual for the rest.


This mission was run with two drones and the Planner assigned one of the sequences to each drone. Figure~\ref{fig:parkour_timeline} depicts a timeline of the Schedulers for both drones.
The drones navigate to the starting positions of their first Shooting Actions and wait for the \texttt{START\_RACE} Event, which triggers both shooting sequences in parallel (only the first Shooting Action of each drone has starting Event associated, the rest are consecutive). Drone 1 approaches to the action scene with a fly-through shot, and then, it performs a flyby in the opposite direction of the runners' movement. Drone 2 starts with a static shot taking an overview of the parkour area. Then, it performs a lateral shot along the scene, which coincides with the runners coming across a complex obstacle. Last, it finishes with a quarter of an orbit around the final part of the scene. 
Some images from the experiment can be seen in Figure~\ref{fig:parkour_mockup}; whereas a complete video is accessible at: \url{https://youtu.be/P_n_PfuEC2A}.

\begin{figure}[tb]
    \centering
    \subfigure[Top view of the scene with RT and drone trajectories for each Shooting Action.]{\label{fig:boat_scheme}\includegraphics[width=1\columnwidth]{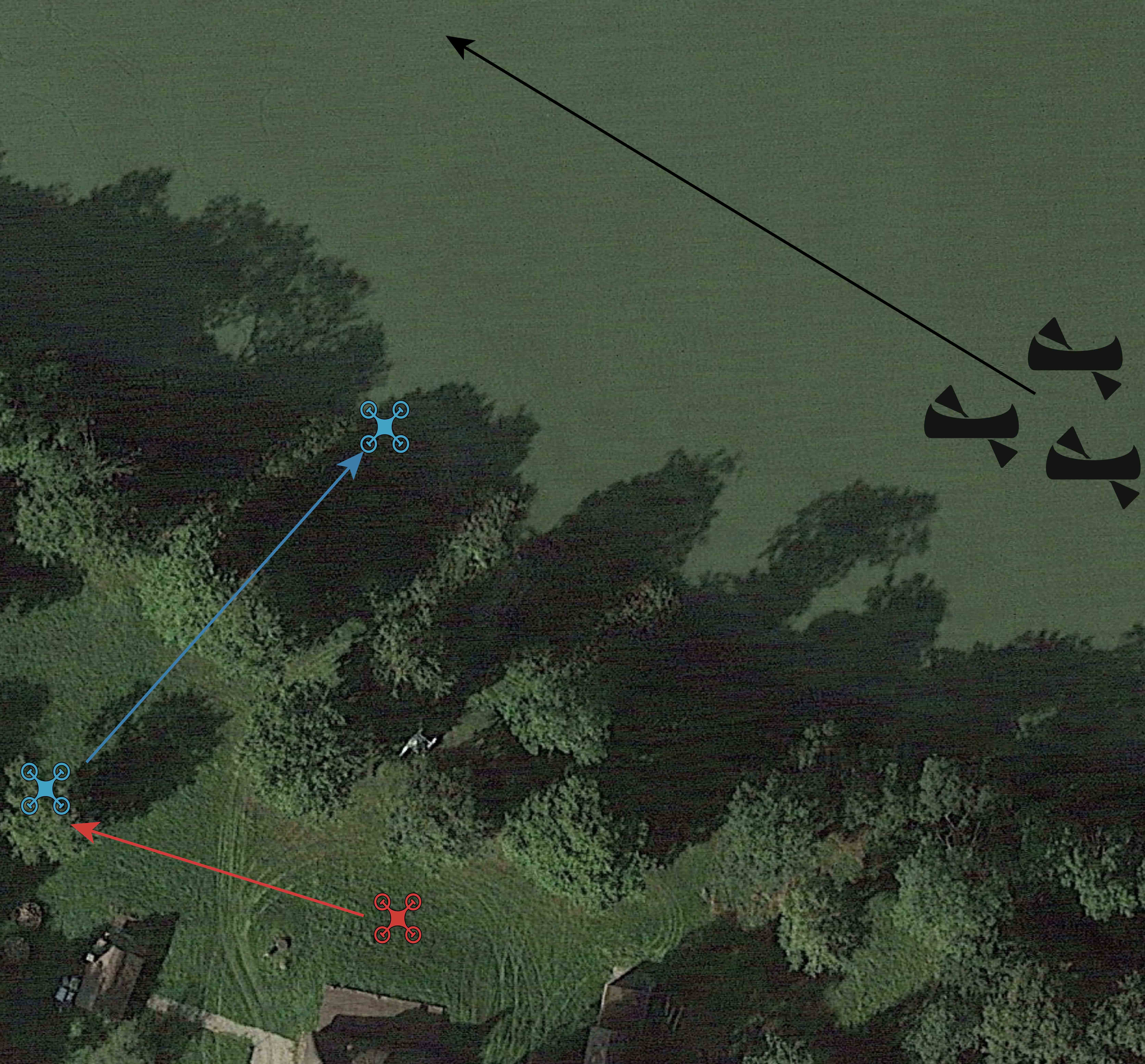}}
    \subfigure[Drone actions over a timeline. Both sequences of consecutive shots are executed in parallel triggered by the \texttt{START\_RACE} Event at time 20~s. A GPS target is tracked.]{\label{fig:boat_timeline}\includegraphics[width=1\columnwidth]{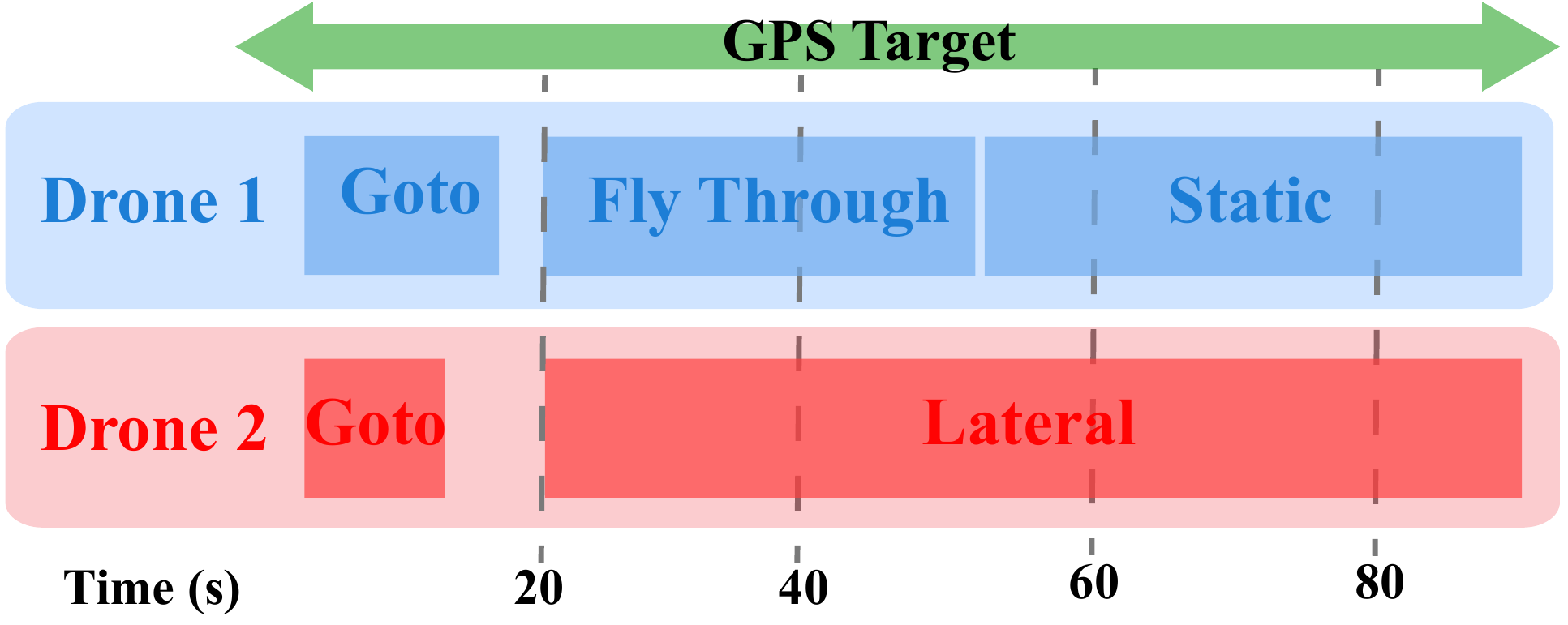}}
    \caption{Rowing race mission with two drones and three different shots. Blue color corresponds to Drone 1 and red color to Drone 2.}
\end{figure}

Additionally, we also demonstrate our system with a mission filming a rowing race. We prepared a mock-up in a lake with four rowing amateur boats recreating a race. Figure~\ref{fig:boat_scheme} shows a scheme of the mission designed by the director. It consists of three shots to film the rowers from the lake's bank as they pass by. A sequence with a fly-through shot followed by a static, and a lateral shot running in parallel; both tracks triggered by the \texttt{START\_RACE} Event. One of the boats carried a GPS target, which was used both as RT and ST for the lateral shot. The fly-through and static shots had none ST.   

\begin{figure}[tb]
    \centering
    \subfigure[Image from camera on board Drone 1 during the static shot.]{\label{fig:boat_1}\includegraphics[width=0.48\columnwidth]{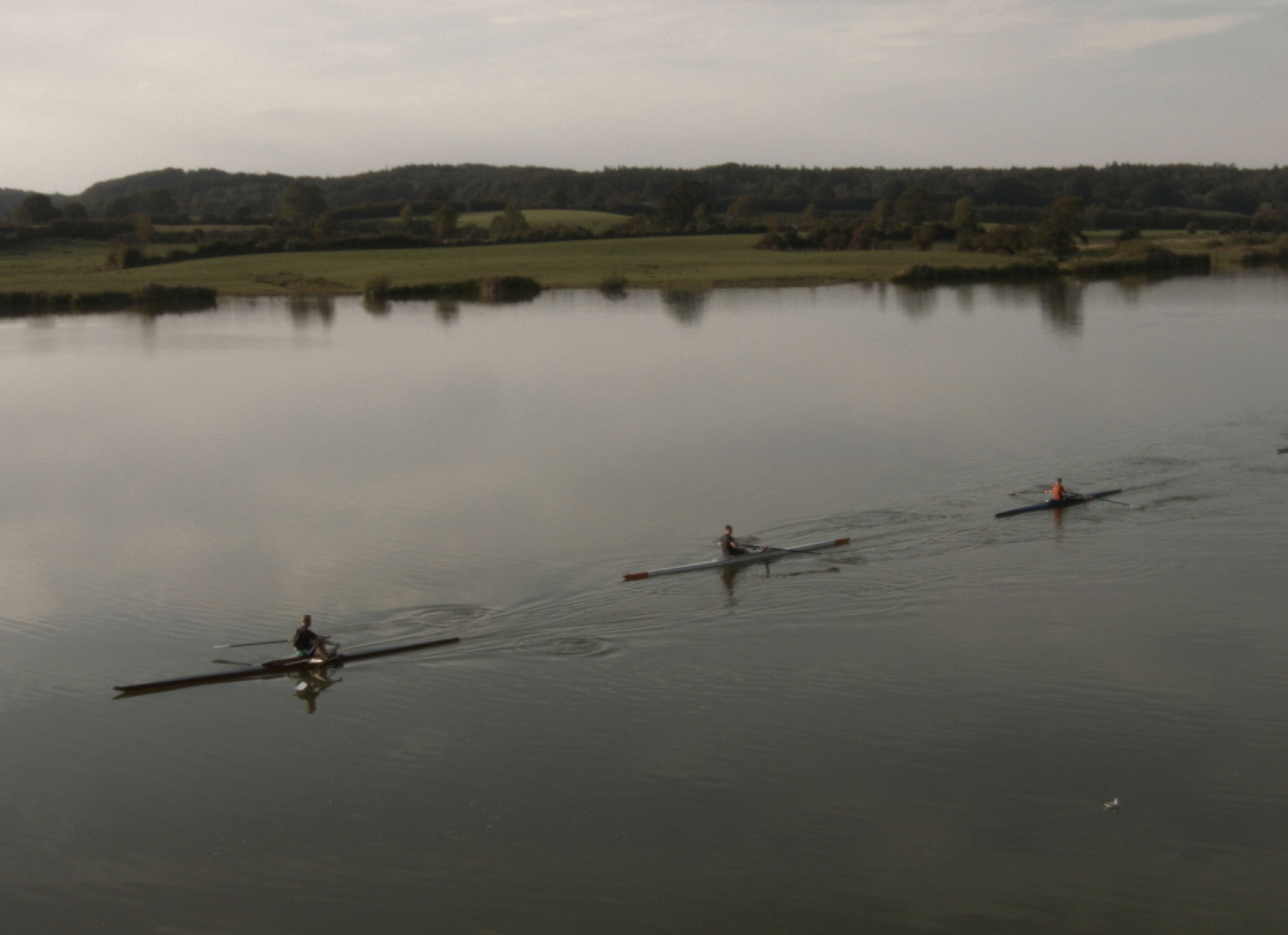}}
    \subfigure[Image from camera on board Drone 2 during the lateral shot.]{ \label{fig:boat_2}\includegraphics[width=0.48\columnwidth]{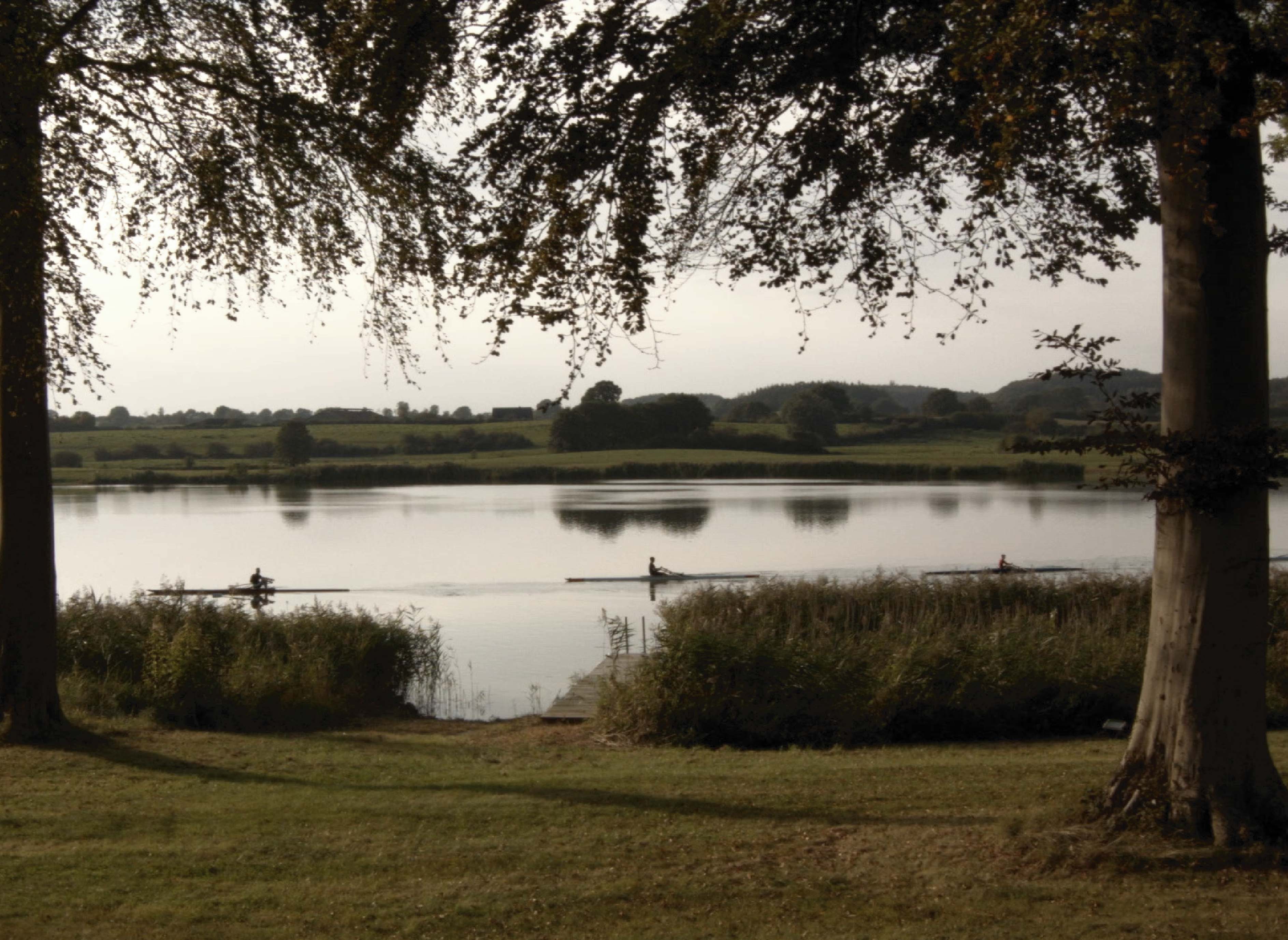}}
    \subfigure[View of the experiment while Drone 1 is taking the static shot as the rowers pass by.]{\label{fig:boat_general_view}\includegraphics[width=1\columnwidth]{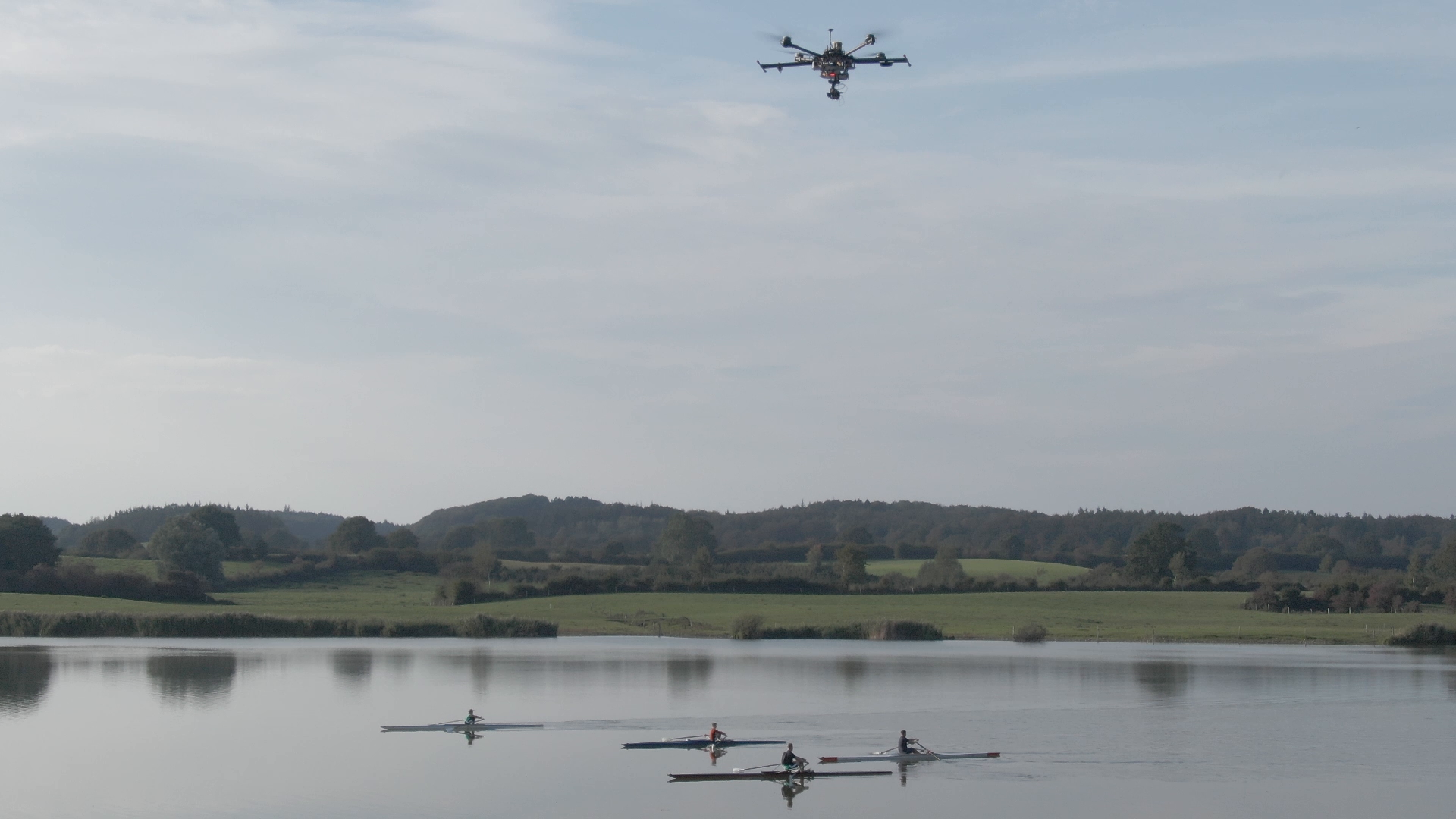}}
    
    

    \caption{Images from a mission with two drones filming a rowing race.}\label{fig:boat_mock_up}
\end{figure}

This mission was run with two drones and the Planner assigned the lateral shot to one drone and the fly-through and the static to the other. Figure~\ref{fig:boat_timeline} depicts a timeline of the Schedulers for both drones.
The drones navigate to the starting positions of their first Shooting Actions and wait for the \texttt{START\_RACE} Event, which triggers both shooting sequences in parallel (only the first Shooting Action of Drone 1 has starting Event associated, the next one happens consecutively). Drone 1 approaches the rowers taking a fly-through shot from the lake's bank over the water. Then, it takes a static shot rising up 10 meters and panning to the left to target the boats. Drone 2 performs a lateral shot over a green area beside the lake's bank, tracking the boats at a 50-meter distance and at a 3-meter height. Some images from the experiment can be seen in Figure~\ref{fig:boat_mock_up}; whereas a complete video is accessible at: \url{https://youtu.be/COay0hZsMzk}.

Finally, we demonstrated the system in a real regatta event in Wannsee, in Berlin (Germany). We deployed the system earlier in the morning of the actual race, in a strategic spot. Then, two of our drones waited for a manually triggered Event to run a short mission designed beforehand, consisting of a static and fly-through shots in parallel, followed by a flyby. The two first shots used none ST, but we employed a visual ST for the last one to track the boats as they passed by. The main objective was just to showcase to media end-users the possibilities of our system and its fast deployment for covering a real sport event. A feature video with the main results of our field campaigns can be seen at \url{https://www.youtube.com/watch?v=iLs6Xo87j78}.

\subsection{Lessons Learned}

In this section, we discuss the main lessons that we learnt during our field and integration tests, about our system and autonomous cinematography with drones in general.

\textit{One drone to rule them all:}
Due to the ambition of the application, the hardware design of the aerial platform was complex. On the one hand, drones were thought to fulfill with safety and usability concerns. This entails the integration of heavy payload, including a high-performance cinematographic camera, several processing units, an LTE module and enough batteries to cover a reasonable flight time (20 minutes). On the other hand, more "commercial" products are usually aimed at smaller platforms, mainly due to logistic and cost constraints. From the feedback of media end-users, we derived that our drones were appropriate to test and demonstrate system functionalities, but a final product should trade off capabilities with payload and size, in order to be more secure and practical for media production.     
 
\textit{Camera and gimbal integration:}
The selection of the camera was quite relevant for the system. A high-performance camera was a requirement from media end-users, and our choice fulfilled with all media specifications. However, we discovered throughout our experimentation, that this kind of cameras are not thought to be integrated in autonomous platforms. First, gimbal calibration was tough, as off-the-shelf gimbals are designed for lighter cameras. A custom product with the camera integrated had been more appropriate to get less shaky images.
Second, we experienced many issues with drivers for video streaming on the NVIDIA TX2, as the selected carrier board (AUVIDEA) did not have official drivers for HDMI input with TX2.
Last, we found problems to focus the camera remotely, which was another requirement from end-users. The camera offered an expansion port to send commands to configure the camera, but this port did not send back any feedback from the camera about its properties (e.g. focus, ISO, white balance, etc). Therefore, it was difficult to implement specific controllers, so we opted for interfacing with the built-in autofocus of the camera, which was not perfect when flying far from the target. 
In general, all these details to integrate commercial cameras and gimbals for high-performance media production on drones are not negligible, and should be considered carefully when designing the system.   

\textit{Middleware and communication:}
We found quite helpful our choice for ROS and UAL as system middleware, as they offered us a good solution to get abstracted from low-level drone control and communication, speeding up software development. UAL also allowed us to design the system transparently, regardless of the final selection of the drone autopilot. In terms of communication, the Thales LTE module provided high-quality video transmission and multi-drone communication, which was critical for the application. However, ROS configuration (we used the \textit{multimaster-fkie} package) to operate with multiple drones in a distributed fashion was troublesome. We believe that the establishment of ROS 2 will be key for multi-robot applications, as communication is decentralized and professional middleware can be easily integrated. 

\textit{Simulation is key in cinematographic applications:}
We used SITL simulations in Gazebo to integrate and test our system, which was tremendously useful to speed up the development process. Nonetheless, simulation turned out to be a helpful tool for media production too. The media director always found interesting to see a 3D recreation of the mission before the real scene happened. For security, we also used the simulator to show graphically to the safety pilots the behavior of the drones before every field test. Even though Gazebo was enough for our purposes, the use of simulators with more realistic graphics engines like \textit{AirSim}~\footnote{https://github.com/microsoft/AirSim.} would be more appealing for media users, enhancing their experience. 

\textit{Media end-users need for alternative types of targets:}
We followed media users' recommendations to implement shots based on both actual and virtual targets. Our way of describing shots by means of a Reference and a Shooting Target, and our different RT modes were a success, as they provided the director with the required level of flexibility. Being able to define virtual rails for camera motion independent of the actual target is highly desirable for media directors. We also learnt that, although they appreciate autonomous functionalities, they also feel the need to have the possibility of operating the gimbal and the focus manually, in order to adapt to the artistic wishes of the director at any moment. 
Regarding shooting targets, we discovered that a combination of GPS and image processing was the best solution. Relying only on GPS can be noisy in locations with adverse GPS conditions, such as near high trees or buildings, but it detects targets with a longer range. A wise trade-off was to use GPS to initially locate the target on the image, and then visual processing to track it more precisely.    

\textit{Onboard collision avoidance is a must:}
Even if the system is to be operated in open and well-structured environments, autonomous collision avoidance is quite relevant, as it provides reassurance to end-users. In this sense, the sooner conflicts are detected, the better. This means that planning components that output solutions minimizing hypothetical conflicts for the drones are desirable. Nonetheless, onboard mechanisms for reactive collision avoidance during mission execution are also necessary to cover dynamic scenes, as there are always unexpected obstacles and inaccuracies in plan execution, mainly for outdoor events. Moreover, relying only on a safety pilot is sometimes tricky as their perspective with respect to the drone is not always ideal. 

\section{Conclusions}
\label{sec:conclusions}

This paper presented a system for autonomous execution of cinematography missions with multiple drones. We introduced the complete architecture, including components for mission design, planning and execution. Then, we focused on the system for mission execution. In particular, we described our parametric manner to define shots, considering different types of camera motion and target actors in the scene. Besides, we implemented a series of canonical shots and proposed a distributed scheduling procedure to execute cinematography missions, which can include sequential and concurrent shots, as well as single- and multi-camera shots. An event-based mechanism is used to synchronize shot execution and to increase the system robustness regarding possible inaccuracies during the planning phase. 

The system was developed within the framework of the EU-funded project MultiDrone and it has been released as open-source for the community. Our field experiments filming sport activities showcased the feasibility of the system to address outdoor cinematography missions involving multiple drones and a variety of shot types. 
As general conclusion, the feedback coming from the media experts in the project was positive, as they found helpful the combination of virtual and actual targets to guide camera motion; as well as the flexibility that our concepts of Reference and Shooting Targets provided.  

As future work, we plan to run more specific subjective user studies to better evaluate the artistic possibilities of the system combining multi-camera shots. Moreover, although we integrated solutions for conflict resolution in the planning components and also for reactive collision avoidance between drones during mission execution, we would like to explore mechanisms more oriented to obstacle avoidance in unstructured environments, using onboard sensors for online mapping.

\section*{Acknowledgement}

We would like to thank all partners from the MultiDrone project, and particularly to Rafael Salmoral, Angel Montes, Vasco Sampaio, Miguel Malaca, Paraskevi Nousi, Iason Karakostas, Thomas Aubourg and Gregoire Guerout, for their support during the field experiments. 



\balance{}

\bibliographystyle{IEEEtran}
\bibliography{multidrone_executer}
\EOD
\end{document}